\documentclass[letterpaper, 10 pt, conference]{ieeeconf}

\IEEEoverridecommandlockouts 
\usepackage[pdftex]{graphicx}
\usepackage{amsmath}
\usepackage{booktabs}

\title{Older Adults' Preferences for Feedback Cadence from an Exercise Coach Robot}

\author{Roshni Kaushik$^{1}$ and Reid Simmons$^{2}$
\thanks{$^{1}$Carnegie Mellon University, Robotics Institute, Pittsburgh, PA, {\tt\small roshnika@andrew.cmu.edu}}%
\thanks{$^{2}$Carnegie Mellon University, Robotics Institute, Pittsburgh, PA, {\tt\small rsimmons@andrew.cmu.edu}}%
}

\begin{document}

\maketitle
\thispagestyle{empty}
\pagestyle{empty}

\begin{abstract}
People can respond to feedback and guidance in different ways, and it is important for robots to personalize their interactions and utilize verbal and nonverbal communication cues. We aim to understand how older adults respond to different cadences of verbal and nonverbal feedback of a robot exercise coach. We conducted an online study of older adults, where participants evaluated videos of the robot giving feedback at different cadences for each modality. The results indicate that changing the cadence of one modality affects the perception of both it and the other modality. We can use the results from this study to better design the frequency of the robot coach's feedback during an exercise session with this population. 
\end{abstract}


\section{Introduction}
As researchers try to create robots that integrate in a variety of domains, they explore how robots can move from controlled factories to more complex environments such as healthcare and education\cite{kyrarini_survey_2021}. In these domains, it is crucial for robots to personalize their interactions, be aware of the contextual nuances surrounding their behavior, and effectively utilize verbal and nonverbal communication cues. Robots that do not personalize their behavior and deploy a one-size-fits-all approach could miss out on opportunities to improve an individual's performance and experience based on their preferences and the contextual situation.


In the domain of exercise, guidance and feedback are especially important, as a personalized coach could provide corrections to ensure that the individual performs the exercises correctly with minimal injury risk, as well as providing motivation and encouragement to promote consistency and enjoyment of the exercise experience. In prior work \cite{kaushik_effects_nodate}, we developed a robotic exercise coach that evaluated the exercises that participants performed in real-time and provided feedback in multiple styles. We found that people performed differently and had different preferences for the styles.

One aspect we did not explore in this previous study was people's preferences for feedback cadence. For an exercise coach, some people may prefer feedback more frequently, and others may feel overwhelmed by too much feedback. People may also prefer different frequencies depending on the feedback modality, such as preferring a greater frequency of nonverbal feedback compared to verbal feedback \cite{bani_use_2011}.

In this work, we specifically focus on older adults (ages 60+) and their preferences for verbal and nonverbal feedback in the domain of exercise. Exercise is extremely important in promoting physical, mental, and emotional well-being in older adults, and it is important to understand how they respond to different feedback levels, so an exercise coach robot could personalize its feedback to those preferences \cite{mora_exercise_2018}. Our main research question is:
\begin{itemize}
    \item What are older adults' preferences for verbal and nonverbal feedback for a robot exercise coach?
\end{itemize}

To explore this question, we use the same exercise coach robot in \cite{kaushik_effects_nodate} that evaluates how well the participants perform exercises and provides verbal and nonverbal feedback while they exercise. We generate videos of the robot responding to exercises with different cadences of \textbf{verbal} and \textbf{nonverbal} feedback, and explore through an online study how older adults' preferences change with the different cadences in an online study. We find support for both main effects (changing the cadence of one modality affects the perception of it) and secondary effects (changing the cadence of one modality affects the perception of the other modality).

\section{Related Work}
Researchers have explored how different frequencies of verbal feedback have an impact on performance and subjective experience during human-human interactions. In a card-sorting task, researchers found that the percentage of responses accompanied by verbal feedback affected the number of trials it took the participants to complete the task \cite{bourne_jr_verbal-reinforcement_1967}. When learning a complex movement task, participants preferred a medium cadence of feedback; too high of a frequency reduced the effectiveness of feedback \cite{niznikowski_effectiveness_2013}. Researchers found that including verbal robot commands was the most effective form of communication while maintaining user trust when the robot was teaching the human \cite{nikolaidis_planning_2018}. Robot verbal feedback in a collaborative task with a human improved task performance and the human's subjective experience \cite{st_clair_how_2015}.

Researchers have also explored how nonverbal feedback affects people. An increase in the frequency of genuine teacher smiles during an interaction improved the likelihood of students correcting mistakes \cite{ergul_case_2023}. A study on the effects of affective human-like and robot-specific behavior showed that these behaviors do impact the perception of the robot and the human's affect \cite{rosenthal-von_der_putten_effects_2018}. Robot motion and mimicry had a significant impact on similarity and closeness in robot-mediated communication \cite{choi_movement_2017}. Combining nonverbal gestures with verbal information can further improve the human experience, as found in \cite{salem_friendly_2011}. Our work explores specifically how changing the frequency of both verbal and nonverbal feedback affects the human's perception of the interaction

\section{Methodology}
\subsection{Exercise Evaluation}

The robot firsts evaluates each repetition of the exercise the participant performs in real-time. We choose bicep curls and lateral raises as the two exercises based on prior work (\cite{kaushik_effects_nodate}), which also includes the technical details of the evaluation.
For each rep, the robot computes two kinds of evaluation information. The speed of a rep can be good, fast, or slow. Additionally, the form of a rep can be: good, low range of motion, high range of motion, or generically bad. This evaluation information is vital for the robot to provide specific feedback on the individual's performance. 



\subsection{Robot Feedback}
Once the robot evaluates a new repetition, it should react in a multi-modal way. We consulted with two domain experts\footnote{Ayotoni Aroyo, ACSM-CPT (Exercise Physiologist and Physical Activity Lead at Emory University'sCognitive Empowerment Program); Gustavo J Almeida, PT, Ph.D. (UT Health San Antonio)} 
to determine what kinds of verbal and nonverbal feedback are appropriate for an exercise coach. From the expert discussions, we design a controller that reacts to the specific feedback outputted by the exercise evaluation. For each modality, we have to answer the questions: (1) what should the feedback look like, and (2) when should the feedback occur?

\subsection{Verbal Feedback}
We define three categories of verbal feedback after discussions with our domain experts:
\begin{enumerate}
\item Positive - $\alpha_1$ good evals or $\alpha_2$ good speed in a row
\item Negative - $\beta_1$ bad evals with the same message or $\beta_2$ fast/slow in a row
\item Correction - $\beta_1$ bad evals with the same message followed by a good eval or $\beta_2$ fast/slow in a row followed by a good speed
\end{enumerate}
where $\alpha_1, \alpha_2, \beta_1, \beta_2$ are all determined by which cadence level is chosen. The robot gives positive or negative feedback generally if it sees a pattern (good or bad) in the human's behavior. We choose to structure the feedback in this way because our domain experts indicated the importance of ensuring that the robot sees a pattern before intervening. Additionally, they emphasized the importance of rewarding a correction of behavior (third bullet), so the human knows that they have successfully incorporated the robot's correction.

We choose values (based on discussions with our experts) for each parameter for \{\emph{low}, \emph{medium}, \emph{high}\} \textbf{verbal} feedback: $\alpha_1 = \{4, 3, 2\}$, $\alpha_2 = \{5, 4, 3\}$, $\beta_1 = \{3, 2, 1\}$, and $\beta_2 = \{4, 3, 2\}$. Note that when the robot has multiple options for how to respond (form and speed verbal feedback), the robot will prioritize form feedback. Additionally, the robot says a message only if it has not finished a previous message 3 seconds prior (experimentally determined) to not overwhelm the human with verbal feedback. The robot has a set of options for exact messages to utter based on the situation. Some examples include:
\begin{itemize}
    \item Last few reps were slow $\rightarrow$ Nice job, can you speed up a little on the next few?
    \item Last few reps had a low range of motion $\rightarrow$ You are doing great, try to get a full range of motion in your elbows.
    \item Last few reps were good form $\rightarrow$ Nice job, looking great!
\end{itemize}


\subsection{Nonverbal Feedback}
The robot's \textbf{nonverbal} feedback has two forms: facial expressions and body movements. Details of the feedback generation can be found in \cite{kaushik_effects_nodate}, and the robot reacts \emph{positively} with a happy facial expression and body movement and \emph{negatively} with a sad facial expression and body movement.



Verbal-nonverbal incongruence is a studied area in human-human interactions, and the interpretation of communication where the nonverbal and verbal reactions do not match can be unclear \cite{creek_responses_1972}. To avoid any potential misinterpretations, we choose to match the nonverbal reactions to the verbal utterances of the robot (\emph{positive} with a positive or correction verbal phrase and \emph{negative} with a negative verbal phrase).

When the robot does not utter anything verbally after a rep, it has a choice of whether or not to react nonverbally. We want to test different frequencies of reacting at these points in the session where the robot is silent. We choose to test three different cadence levels of the \textbf{nonverbal} feedback. For the \emph{high} cadence, the robot always reacts positively or negatively; the robot reacts only 50\% of the time for the \emph{medium} cadence; and it reacts only 25\% of the time for the \emph{low} cadence. We choose these numerical values as they were visually distinguishable when tested on the robot and represent a large variety of response frequencies.



\vspace{-0.4em}
\section{Study Design}
Our user study tests the preferences of older adults for different levels of \textbf{verbal} and \textbf{nonverbal} feedback. After discussions with our domain experts, we choose to test the cadence levels of \textbf{nonverbal} feedback at or above the cadence levels of \textbf{verbal} feedback to reduce the number of combinations of cadences to test. For example, when the \textbf{verbal} cadence is high and the \textbf{nonverbal} cadence is low, the robot would react verbally very frequently, so changes in the \textbf{nonverbal} cadence (which only change after the reps where the robot does not react verbally) would be minimal. However, when the \textbf{verbal} cadence is low, there are many opportunities for the \textbf{nonverbal} cadence to affect the number of nonverbal reactions the human sees since there are many reps to which the robot does not react verbally. 

\begin{figure}[ht]
    \centering
    \includegraphics[width=0.6\columnwidth]{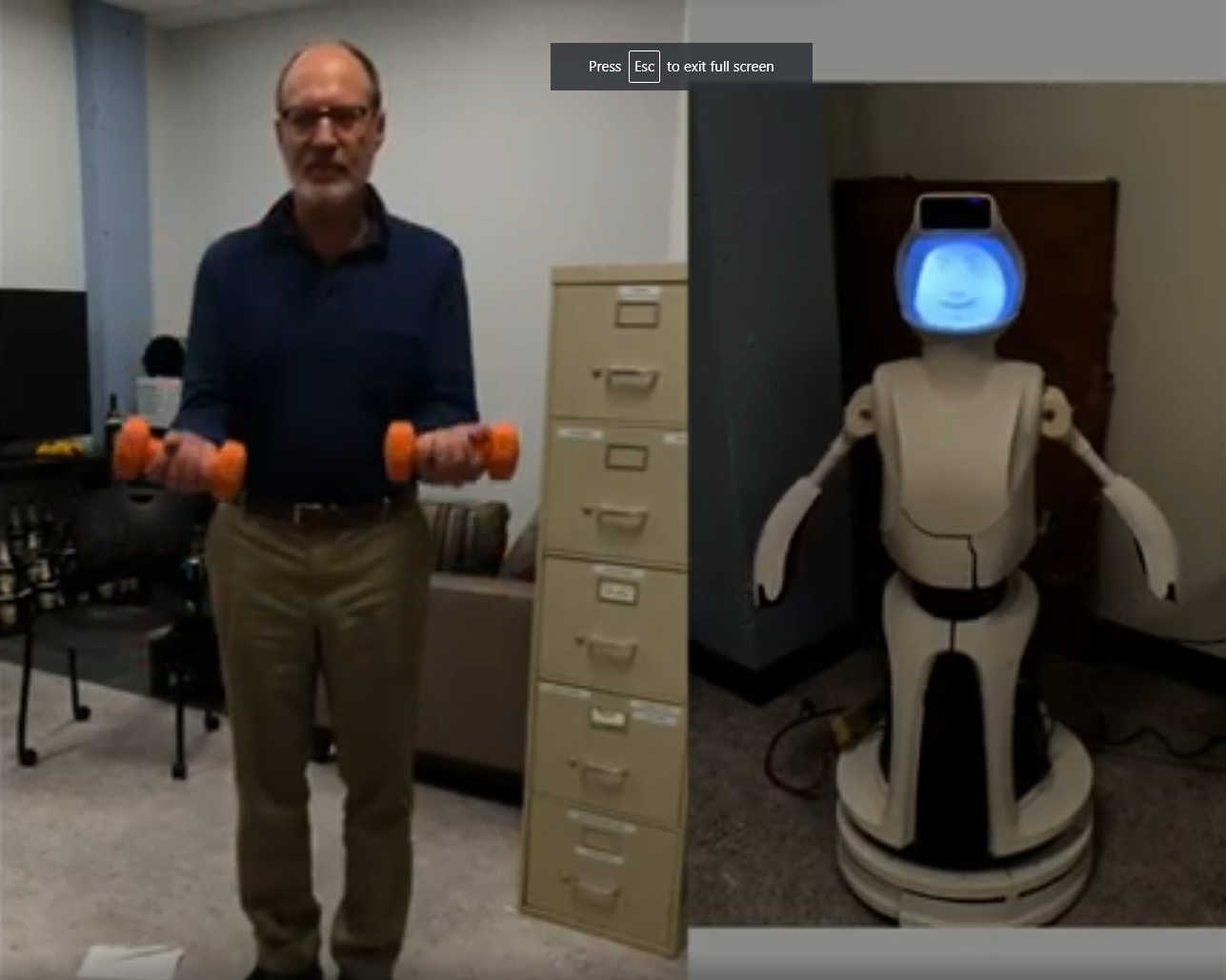}
    \caption{Screenshot for a video from the study with the human exerciser on the left and the robot providing verbal and nonverbal feedback on the right.}
    \label{fig:video_screenshot}
\end{figure}

For each of the 6 conditions, we record one session of bicep curls and one session of lateral raises. Each session includes: 4 good form/good speed, 3 low range form/good speed, 3 good form/slow, and 5 good form/good speed. This results in a total of 12 videos (see Figure \ref{fig:video_screenshot} for screenshot), 2 per condition.

Participants begin this online study on Qualitrics with a consent form and demographic information. They read through an explanation of the videos they will see in the study that includes a description of the two feedback modalities. They also complete one training question that asks ``Which of the following will the robot change as part of its feedback?'' where the required answer is selecting facial expressions, body movements, and what the robot says. This question is intended to ensure the participants pay attention to the robot's behavior.

They then see one video per condition (randomly chosen between the two videos in each condition), with the order of the videos randomized. After seeing each video, the participants complete two sets of questions: one about their impressions of the robot's \textbf{verbal} feedback and one about the \textbf{nonverbal} feedback. On a 5-pt scale from strongly disagree to strongly agree, they evaluate the usefulness, clarity, timeliness, and helpfulness of each type of feedback. They then have the option to explain their choice in a free-form text format.

We discuss the results and hypotheses representing main effects (changes in a modality affect the perception of it) and secondary effects (changes in a modality affect the perception of the other modality) in the following section.

\begin{figure*}[ht]
    \centering
    \includegraphics[width=0.8\textwidth]{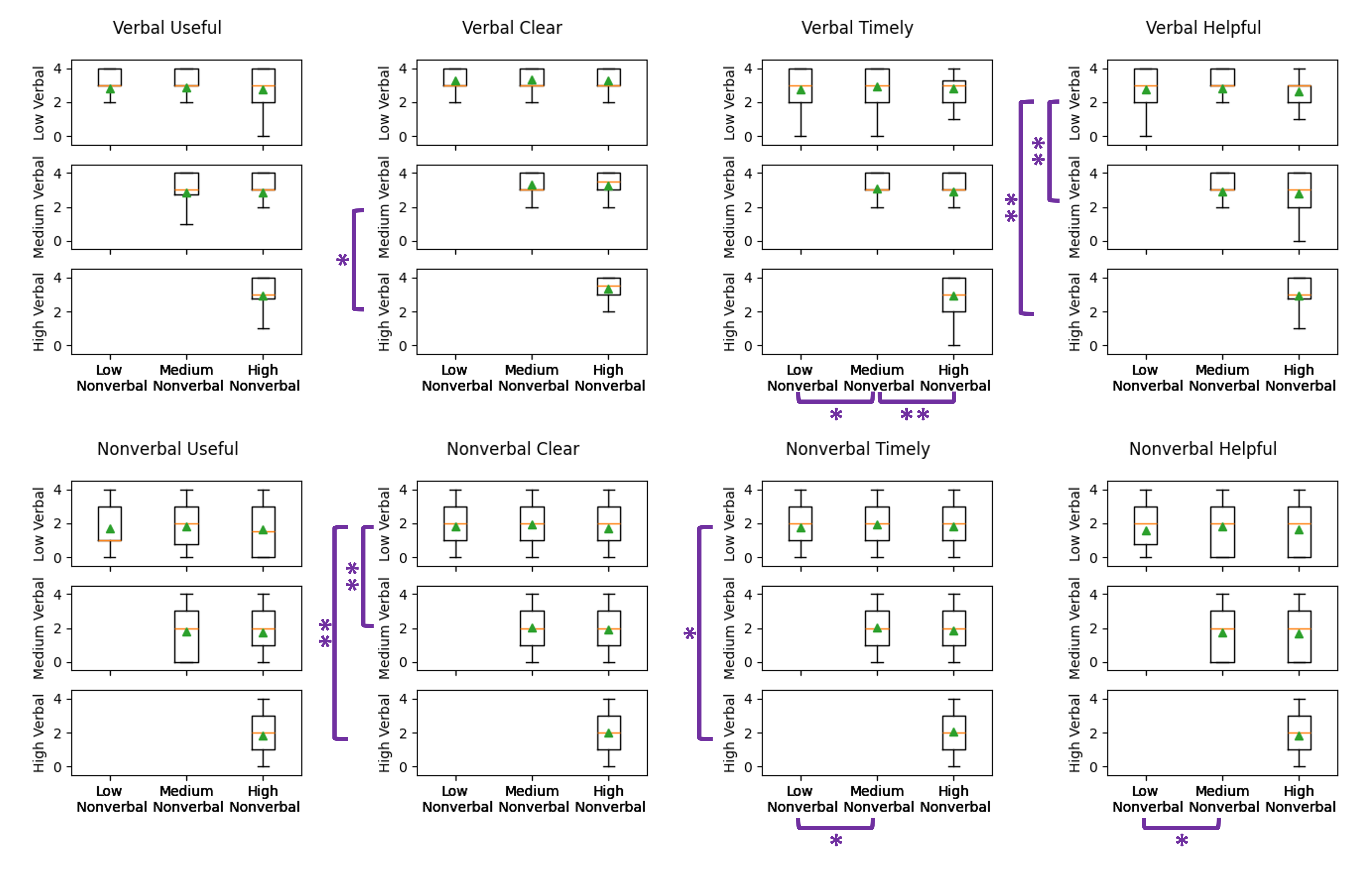}
    \caption{Results of all Likert-style measures with perceptions of verbal feedback in the first row and nonverbal feedback in the second row. Significant differences are marked with one asterisk ($p<0.05$) or two asterisks ($p<0.01$). Each box plot shows the interquartile range of the data, as well as the min and max of the data. The mean is shown with a green triangle.}
    \label{fig:results}
    \vspace{-1.5em}
\end{figure*}


\vspace{-0.2em}
\section{Results and Discussion}
We recruited 100 online participants using Prolific, with a criterion of age greater than 60, as we want to focus on older adults' perceptions of the robot, with the study protocol approved by the IRB. We convert the Likert-style results into numeric values from 1-5 and perform a repeated measures ANOVA with the \textbf{verbal} and \textbf{nonverbal} cadences as two within-subjects conditions, using the Greenhouse-Geisser corrected p-value. Where we found significance, we then perform a pairwise post-hoc test to determine which pairwise differences between feedback cadence levels are significant. Figure \ref{fig:results} shows the results marked with significant differences.

\subsection*{\textbf{H1:} Changing the \textbf{verbal} feedback cadence affects the participants' view of the \textbf{verbal} feedback}
We have support for this hypothesis. Changing the \textbf{verbal} feedback cadence has an effect on the perceived clarity and helpfulness of the robot's \textbf{verbal} feedback. Specifically, participants seemed to prefer the \emph{high} level compared to the \emph{medium} in terms of clarity ($p<0.05, F(2, 198)=3.26$) and the \emph{medium} and \emph{high} over the \emph{low} in terms of helpfulness ($p<0.01, F(2,198)=6.47$). This result illustrates that the \emph{low} level of verbal feedback is generally not preferred and that \emph{medium} and \emph{high} levels are perceived as more helpful and clear. The more frequent feedback could make the robot feel more responsive and aware of what the human is doing.

\subsection*{\textbf{H2:} Changing the \textbf{nonverbal} feedback cadence affects the participants' view of the \textbf{nonverbal} feedback}
We have support for this hypothesis. Changing the \textbf{nonverbal} feedback cadence has an effect on the perceived timeliness ($p<0.05, F(2, 198)=3.70$) and helpfulness ($p<0.05, F(2,198)=3.42$) of the robot's feedback. Participants preferred a \emph{medium} level for both of these measures over the \emph{low} level. The higher frequency of nonverbal reactions could be more engaging for participants, and they could feel that the robot is actually responding to the exercises.

\subsection*{\textbf{H3:} Changing the \textbf{verbal} feedback cadence affects the participants' view of the \textbf{nonverbal} feedback}
We have support for this hypothesis. Changing the robot's \textbf{verbal} feedback cadence affects the perception of the \textbf{nonverbal} feedback's clarity ($p<0.01, F(2, 198)=4.39$) and timeliness ($p<0.05, F(2, 198)=3.75$). Participants thought the \emph{medium} and \emph{high} levels of verbal feedback paired with better clarity and timeliness of the nonverbal feedback. This is an interesting secondary effect that shows that the two feedback modalities are inexorably linked. Changing the cadence of one will affect the perception of the other.

Psychology researchers have explored this interaction, and \cite{lovaas_interaction_1961} explores many possibilities for how they are linked, including aggressive verbal utterances signaling aggressive nonverbal reactions. In our results, changing the verbal frequency could help participants interpret the nonverbal feedback better and make it appear more clear and timely.

\subsection*{\textbf{H4:} Changing the \textbf{nonverbal} feedback cadence affects the participants' view of the \textbf{verbal} feedback}
We have support for this hypothesis. Changing the robot's \textbf{nonverbal} feedback cadence affects the perception of the \textbf{verbal} feedback's timeliness ($p<0.05, F(2, 198)=4.36$). Participants preferred the \emph{medium} level of nonverbal feedback when thinking about the timeliness of verbal feedback. This also agrees with our finding in \textbf{H2} that the \emph{medium} nonverbal cadence is preferred to the \emph{low} level.



\section{Conclusion}
We present a robot exercise coach with an exercise evaluator that provides both verbal and nonverbal feedback. We then introduce an online user study that explores how older adults perceive different cadence levels of the two feedback modalities by showing them videos with different combinations of feedback levels. The results support all our hypotheses for both main effects (changing the cadence of one modality affects the perception of it) and secondary effects (changing the cadence of one modality affects the perception of the other modality). Our results allow us to choose the feedback cadences that older adults prefer for this robot exercise coach.

The main limitation of our work is that the results are based on participants viewing a video of the robot giving feedback to another individual rather than experiencing the feedback themselves. People may have different perceptions of feedback when directed at themselves, so we will keep that in mind when generalizing these results to in-person interactions. Additionally, when people view the robot giving multi-modal feedback, they may not be fully aware of how it is affecting them. Qualitative comments in prior work indicated that some people were unaware of the effect of nonverbal feedback on their performance \cite{kaushik_affective_2022}. This highlights the differences between actual and perceived impact of feedback.

In this work, we do show that older adults have preferences for specific levels of verbal and nonverbal feedback, and we can use these results to better design the frequency of the robot's feedback during the exercise coach session. 

\addtolength{\textheight}{-12cm}

\section*{ACKNOWLEDGMENT}
This work was funded by NSF Grant \#2112633.



\bibliographystyle{IEEEtran}
\bibliography{bibliography}

\end{document}